\documentclass[conference]{IEEEtran}
\IEEEoverridecommandlockouts

\usepackage{cite}
\usepackage{amsmath,amssymb,amsfonts}
\usepackage{algorithmic}
\usepackage{graphicx}
\usepackage{textcomp}
\usepackage{hyperref}
\usepackage{booktabs}
\usepackage{multirow}
\usepackage{caption}
\usepackage{subfigure}
\usepackage{tcolorbox}
\tcbuselibrary{listingsutf8}
\newtcolorbox[auto counter]{promptbox}[2][]{%
  colback=gray!5!white, 
  colframe=gray!75!black, 
  fonttitle=\bfseries,
  fontupper=\footnotesize,
  title=Prompt~\thetcbcounter: #2, 
  label={#1}
}

\def\BibTeX{{\rm B\kern-.05em{\sc i\kern-.025em b}\kern-.08em
    T\kern-.1667em\lower.7ex\hbox{E}\kern-.125emX}}
\begin{document}

% TITLE
\title{A Transformer-Based Cross-Platform Analysis of Public Discourse on the 15-Minute City Paradigm}

% AUTHORS
\author{\IEEEauthorblockN{
Gaurab Chhetri\IEEEauthorrefmark{1}, 
Darrell Anderson\IEEEauthorrefmark{2}, 
Boniphace Kutela, Ph.D.\IEEEauthorrefmark{3}, 
Subasish Das, Ph.D.\IEEEauthorrefmark{2}
}
\IEEEauthorblockA{\IEEEauthorrefmark{1}College of Science and Engineering, Texas State University, San Marcos, Texas, USA\\
Email: gaurab@txstate.edu}
\IEEEauthorblockA{\IEEEauthorrefmark{2}Ingram School of Engineering, Texas State University, San Marcos, Texas, USA\\
Email: \{dma147, subasish\}@txstate.edu}
\IEEEauthorblockA{\IEEEauthorrefmark{3}Texas A\&M Transportation Institute, Texas A\&M University, Houston, Texas, USA\\
Email: b-kutela@tti.tamu.edu}
}

\maketitle

\begin{abstract}
This study presents the first multi-platform sentiment analysis of public opinion on the 15-minute city concept across Twitter, Reddit, and news media. Using compressed transformer models and Llama-3-8B for annotation, we classify sentiment across heterogeneous text domains. Our pipeline handles long-form and short-form text, supports consistent annotation, and enables reproducible evaluation. We benchmark five models (DistilRoBERTa, DistilBERT, MiniLM, ELECTRA, TinyBERT) using stratified 5-fold cross-validation, reporting F1-score, AUC, and training time. DistilRoBERTa achieved the highest F1 (0.8292), TinyBERT the best efficiency, and MiniLM the best cross-platform consistency. Results show News data yields inflated performance due to class imbalance, Reddit suffers from summarization loss, and Twitter offers moderate challenge. Compressed models perform competitively, challenging assumptions that larger models are necessary. We identify platform-specific trade-offs and propose directions for scalable, real-world sentiment classification in urban planning discourse.
\end{abstract}

\begin{IEEEkeywords}
15-minute cities, walkable cities, urban life, sentiment analysis.
\end{IEEEkeywords}

\section{Introduction}

The 15-minute city, introduced by Carlos Moreno, advocates for localized, sustainable urban living by ensuring essential services—such as groceries, healthcare, schools, parks, and public transit—are accessible within a 15-minute walk or bike ride \cite{stanley2015connecting, itdp2018pedestrians, Das2024m}. While the model has gained traction globally for its potential to enhance livability and reduce emissions, concerns persist around personal mobility restrictions and the feasibility of retrofitting existing urban layouts. Nevertheless, its promise in promoting equity and bridging urban–suburban gaps remains strong \cite{abdelfattah2022fifteen}. Scholars have studied this model across diverse contexts. In Melbourne, studies highlighted density, transit, and accessibility as key enablers \cite{stanley2015connecting, pwc2017melbourne}. Others have evaluated walkability using tools like Walk Score and GIS \cite{turon2017concept, imam2019walkable}, while spatial planning and user-informed design strategies were explored in Colombia and Japan \cite{gonzalez2020designing, ito2021strategic}. Work from Milan, Naples, Tokyo, and Vancouver emphasized accessibility and equity challenges \cite{abdelfattah2022fifteen, gaglione2022urban, shimizu2022realizing, hosford2022fifteen}, and additional studies addressed public awareness, mapping technologies, and demographic disparities \cite{abdullah2022awareness, logan2022xminute, willberg2023fifteen, zhang2023creating}.

As the concept gained visibility, it also triggered polarized online debates. Platforms like Twitter and Reddit became forums for both advocacy and opposition, shaped by misinformation, political framing, and skepticism about urban control \cite{Das2019_ExtractingPatternsFromTwitter, DasDutta2020_CharacterizingPublicEmotions, Das2021a, bruns2013twitter}. This environment complicates traditional sentiment analysis methods and calls for scalable, adaptable tools. Transformer-based natural language processing (NLP) models offer powerful solutions for extracting sentiment from unstructured text, but most prior research relies on single-platform data and large, resource-intensive models. The role of compact transformers in efficient, cross-domain sentiment analysis for contested urban planning topics remains underexplored.

This study benchmarks five lightweight transformer models (DistilRoBERTa, DistilBERT, MiniLM, ELECTRA-small, and TinyBERT) on sentiment classification related to the 15-minute city across Twitter, Reddit, and news articles. Using a dataset of over 120,000 entries annotated via a two-stage Llama-3-8B pipeline, we evaluate each model under consistent training conditions using stratified 5-fold cross-validation. Metrics include F1-score, AUC, and training time, enabling a comprehensive comparison of performance and efficiency. Our findings address two core research questions: the effectiveness of compressed models across platforms with varied linguistic styles, and the trade-offs between classification accuracy and computational cost. The study contributes to both methodological research and practical urban policy analysis by demonstrating scalable tools for monitoring public discourse on sustainability and city design.

\section{Problem Statement}

We formulate sentiment classification on the 15-minute city as a binary text classification task across multiple platforms. Given a dataset $D = {(t_i, s_i, p_i, y_i)}{i=1}^n$, where $t_i$ is the $i$-th text, $s_i$ the source (Twitter, Reddit, or News), $p_i$ the platform metadata, and $y_i \in {0,1}$ the sentiment label (1 = supportive, 0 = opposed), our goal is to learn a function $f\theta: \mathcal{X} \rightarrow {0,1}$ that maximizes the cross-validated F1-score:

\begin{equation}
f^* = \arg\max_{f_\theta} \text{F1-Score}{\text{CV}}(f\theta, D)
\end{equation}

F1-score is computed as:

\begin{equation}
\label{eq:f1-score}
\text{F1-Score} = 2 \cdot \frac{\text{Precision} \cdot \text{Recall}}{\text{Precision} + \text{Recall}}
\end{equation}

We apply stratified 5-fold cross-validation to address label imbalance and ensure robust performance estimation.

\section{Methodology}

\begin{figure}[ht]
\centering
\includegraphics[width=1\linewidth]{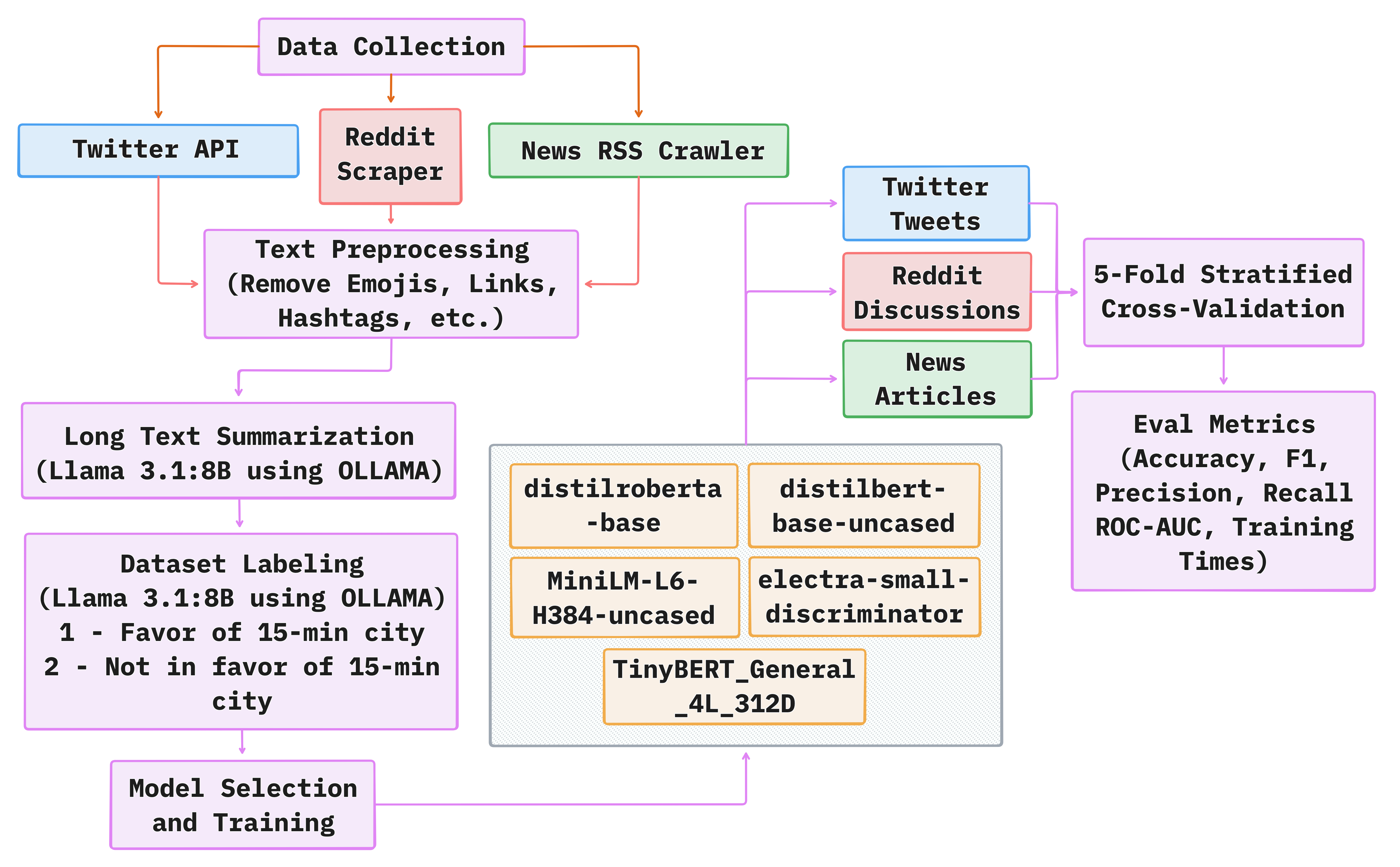}
\caption{Study overview: data collection, summarization, LLM-based labeling, and model evaluation.}
\label{fig:study-design}
\end{figure}

This study implements a reproducible pipeline for collecting multi-platform data, applying LLM-based summarization and labeling, and evaluating compressed transformer models under uniform training. It supports diverse text types and includes four stages: data collection, annotation, 5-fold classification, and performance evaluation. See ~\autoref{fig:study-design} for an overview.

\subsection{Data Collection and Annotation}

We developed a comprehensive multi-platform data collection framework to capture diverse opinions on the 15-minute city concept across three media environments. Each platform's unique characteristics required tailored strategies for effective data acquisition and processing.

\subsubsection{Twitter Data Collection}

We utilized historical Twitter API access during the period of full functionality to retrieve tweets containing keywords related to the 15-minute city. A keyword expansion strategy, informed by prior terminology analyses \cite{UoB_15minute_cities_2024, Steuteville_2023, FerretJournalists_2023}, included direct terms ("15 minute city", "15-minute city", "fifteen minute city"), related ideas ("walkable neighborhoods", "car-free cities"), and policy phrases ("traffic restrictions", "low traffic neighborhoods", "active mobility"). The dataset spans from January 1, 2016, to May 30, 2023, and contains 115,248 tweets after filtering out retweets to preserve original commentary. Each tweet record includes 31 metadata fields such as 'tweet id,' 'text,' 'timestamp,' and 'sourcetweetid.'

\subsubsection{Reddit Data Collection}

We employed a custom scraping framework\footnote{\url{https://github.com/gauravfs-14/reddit-scraper}} (MIT License) targeting ten subreddits spanning urban planning, transportation, sustainability, and diverse ideological perspectives. We scraped posts from 10 subreddits spanning urban planning, sustainability, and ideological perspectives (e.g., r/urbanplanning, r/walkablecities, r/conspiracy). The framework captured both original posts and full comment threads, preserving hierarchical discussion structure and thread context. Reddit posed challenges due to long-form discourse, with some threads spanning thousands of words across multiple nested levels.

\subsubsection{New Media Data Collection}

An automated extraction tool\footnote{\url{https://github.com/gauravfs-14/gnews-collector-rss}} (MIT License) was developed to collect full news articles instead of relying on headlines or summaries, enabling deeper analysis of editorial framing and content.

The final dataset comprised 4897 Reddit discussions, 427 news articles, and 115,248 Twitter posts, all subsequently labeled using our LLM-based annotation pipeline described in \autoref{sec:labelining-pipe}.

\subsection{Model Architecture}

We benchmark five compressed transformer models that balance performance and efficiency across diverse text domains. \textbf{DistilRoBERTa-base} is a distilled version of RoBERTa with 6 transformer layers, 768-dimensional embeddings, and 82M parameters (vs. 125M in the original), achieving near-parity in performance with approximately 2× faster inference \cite{Sanh2019DistilBERTAD}. \textbf{DistilBERT-base-uncased} similarly reduces BERT’s size by 40\% while retaining over 95\% of its accuracy, and offers 60\% faster inference—making it suitable for large-scale applications \cite{Sanh2019DistilBERTAD}. \textbf{MiniLM-L6-H384-uncased} uses 6 layers and 384-dimensional hidden states, preserving essential language understanding by retaining every other layer from its 12-layer parent, and is optimized for low-resource environments \cite{wang2020minilm}. \textbf{ELECTRA-small-discriminator}, with 14M parameters and 256-dimensional hidden states, employs a replaced-token detection pretraining objective that improves sample efficiency over masked language modeling and enables fast, effective fine-tuning \cite{clark2020electra}. Lastly, \textbf{TinyBERT\_General\_4L\_312D} is an ultra-compact model with 4 layers and 312-dimensional embeddings, trained via general and task-specific distillation, achieving a 7.5× size reduction and 9.4× speedup over BERT-base while maintaining competitive performance for real-time or edge deployment \cite{jiao2019tinybert}.

\subsection{Training and Evaluation Framework}

\subsubsection{Automated Annotation Pipeline}
\label{sec:labelining-pipe}
We implement an automated annotation pipeline using Meta's Llama-3-8B \cite{dubey2024llama} model accessed locally via Ollama. The annotation process first summarizes long texts into around 2 sentences using the same model using the prompt~\ref{box:summarization-prompt}, and following the summarization, the annotation process employs carefully designed prompts that explicitly instruct the model to classify content based on support or opposition to the 15-minute city concept, returning binary labels (1 for supportive, 0 for opposed) without explanatory text using the prompt~\ref{box:labeling-prompt}.

\begin{promptbox}[box:summarization-prompt]{Summarization}
You are a smart summarization assistant. Summarize the following social media or news text into a short, focused version that retains only the key opinion, sentiment, or stance related to \textbf{15 minute city} and \textbf{walkable city} topics. If the text is not related to 15 minute city or walkable city, return the summary of the main topic as discussed in the text.

\medskip
Your summary should:
\begin{itemize}
    \item Be no longer than 2 sentences.
    \item Preserve the speaker's opinion or sentiment toward the 15 minute city concept.
    \item Omit unrelated context or background.
\end{itemize}

Return only the summary without extra explanation or any text like \texttt{"Summary:"} or \texttt{"Here is the summary:"}.

\medskip
\textbf{Original text}
\begin{quote}
\textit{text}
\end{quote}
\end{promptbox}

\begin{promptbox}[box:labeling-prompt]{Classification}
You are a data annotation assistant. Your task is to determine whether the following text from social media or news is in favor of \textbf{15 minute city, walkable city} or not.

Label the text as:
\begin{itemize}
    \item 1 if it is in favor of 15 minute city, walkable city.
    \item 0 if it is not in favor of 15 minute city, walkable city.
\end{itemize}

Respond with a \textbf{single digit (0 or 1)}. Do not include any explanation or extra text.

\textbf{Text to analyze:}
\begin{quote}
\textit{text}
\end{quote}
\end{promptbox}

\subsubsection{Training and Evaluation Protocol}

All models were fine-tuned under a unified configuration to ensure fair comparison across platforms and architectures. We used the AdamW optimizer with a fixed learning rate of $2\times10^{-5}$, a batch size of 8 (training) and 16 (evaluation), and trained for up to five epochs with early stopping (patience = 1). Each model employed its native tokenizer with input truncated at 512 tokens. Experiments were conducted on a CPU-only macOS system (Apple M4 Pro, 24 GB unified memory), with mixed-precision training disabled for compatibility. Only the best checkpoint per fold was retained.

Evaluation followed stratified 5-fold cross-validation per dataset (Twitter, Reddit, News), preserving label balance. For each model–dataset pair, performance was averaged across five folds using accuracy, F1-score, precision, recall, AUC, and training time. ROC curves were generated and aggregated per fold to assess discriminative power. This framework supports reproducibility and cross-domain experimentation (e.g., training on Reddit, testing on Twitter). Code, datasets, and scripts are available on GitHub \footnote{\url{https://github.com/gauravfs-14/15min-city-bench}}.

\section{Results}

We conducted a comprehensive evaluation of transformer-based sentiment classifiers across three digital platforms to assess their effectiveness in analyzing public discourse on 15-minute city concepts. The study tested 15 model-platform combinations (5 models × 3 platforms) using stratified 5-fold cross-validation, enabling robust comparisons that account for both architectural differences and platform-specific challenges.

\subsection{Overall Cross-Platform Performance Analysis}

DistilRoBERTa achieved the highest average F1-score (0.8292), followed closely by DistilBERT (0.8281) and MiniLM (0.8276). The minimal performance gap between these top models (all $<$ 0.002) suggests that architectural differences may be less impactful than the training setup. \autoref{tab:merged-overall-efficiency} summarizes detailed metrics across all five models.

TinyBERT stood out for efficiency, posting the fastest average training time (233.22 seconds) while maintaining competitive performance (F1: 0.8242) and the highest recall (0.8888) across platforms. This challenges assumptions that aggressive compression degrades classification quality, demonstrating that a 7.5× parameter reduction relative to BERT-base still preserves strong utility for sentiment analysis.

\begin{figure*}
    \centering
    \includegraphics[width=1\linewidth]{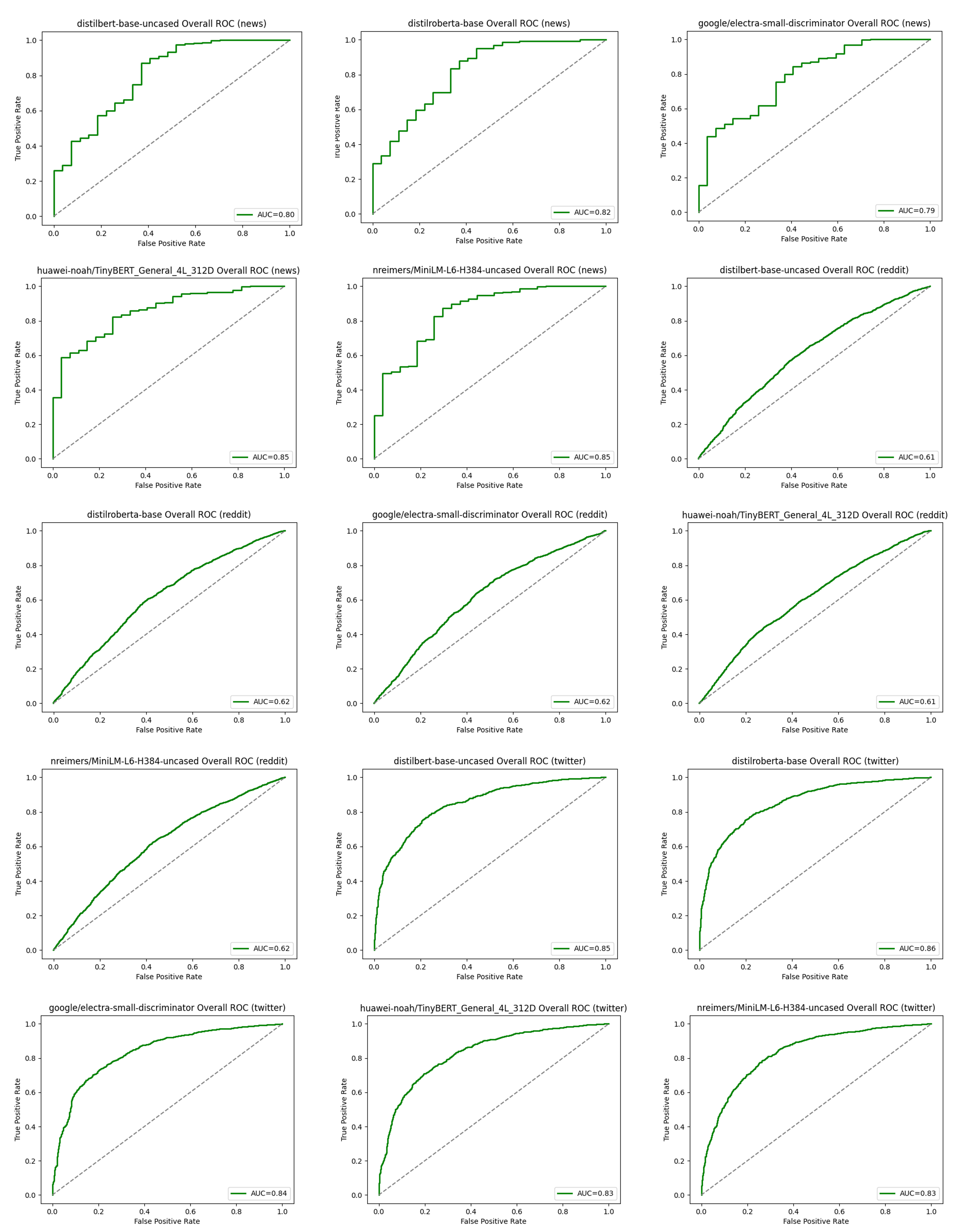}
    \caption{ROC curves for all model–dataset combinations. (Each subplot shows the 5-fold averaged performance for one model on one dataset.)}
    \label{fig:roc}
\end{figure*}

As shown in \autoref{fig:roc}, DistilRoBERTa and MiniLM achieved the highest AUCs on Twitter. Greater performance variance was observed on Reddit and News, but DistilRoBERTa remained consistently strong across all domains.

\begin{table*}[ht]
\centering
\caption{Overall Model Performance and Efficiency Summary}
\label{tab:merged-overall-efficiency}
\small
\begin{tabular}{lcccccc}
\toprule
\textbf{Model} & \textbf{Accuracy} & \textbf{F1-Score} & \textbf{AUC} & \textbf{Training Time (s)} & \textbf{Efficiency} & \textbf{F1 Gap} \\
\midrule
DistilRoBERTa & 0.7759 & \textbf{0.8292} & 0.7642 & 612.63 & 1.353 & 0.2566 \\
DistilBERT    & 0.7751 & 0.8281 & 0.7532 & 528.18 & 1.568 & 0.2588 \\
MiniLM        & 0.7738 & 0.8276 & \textbf{0.7664} & 717.10 & 1.154 & \textbf{0.2560} \\
TinyBERT      & 0.7658 & 0.8242 & 0.7638 & \textbf{233.22} & \textbf{3.535} & 0.2632 \\
ELECTRA       & 0.7723 & 0.8225 & 0.7474 & 371.08 & 2.216 & 0.2570 \\
\bottomrule
\end{tabular}
\\
\vspace{4pt}
\textit{Note: Efficiency = F1/sec $\times$ 1000, and F1 Gap = Best - Worst}
\end{table*}

\begin{table*}[ht]
\centering
\caption{Model Performance by Platform (5-Fold CV)}
\label{tab:merged-platform-performance}
\small
\begin{tabular}{llcccccc}
\toprule
\textbf{Model} & \textbf{Platform} & \textbf{Accuracy} & \textbf{F1-Score} & \textbf{Precision} & \textbf{Recall} & \textbf{AUC} & \textbf{Training Time (s)} \\
\midrule
DistilRoBERTa    & News    & \textbf{0.9413} & \textbf{0.9696} & \textbf{0.9411} & 1.000  & 0.8179 & 67.54 \\
DistilRoBERTa    & Reddit  & 0.6184 & 0.7130 & 0.6448 & 0.7983 & 0.6185 & 1622.35 \\
DistilRoBERTa    & Twitter & 0.7678 & 0.8048 & 0.7777 & 0.8372 & \textbf{0.8563} & 148.00 \\
\midrule
DistilBERT       & News    & 0.9366 & 0.9673 & 0.9366 & 1.000  & 0.8015 & 41.24 \\
DistilBERT       & Reddit  & 0.6144 & 0.7085 & 0.6449 & 0.7922 & 0.6110 & 1435.30 \\
DistilBERT       & Twitter & \textbf{0.7741} & \textbf{0.8084} & \textbf{0.7886} & 0.8333 & 0.8470 & 107.99 \\
\midrule
MiniLM           & News    & 0.9366 & 0.9673 & 0.9366 & 1.000  & 0.8479 & 32.45 \\
MiniLM           & Reddit  & 0.6184 & 0.7113 & 0.6471 & 0.7934 & 0.6186 & 1150.86 \\
MiniLM           & Twitter & 0.7664 & 0.8042 & 0.7729 & \textbf{0.8391} & 0.8328 & 68.00 \\
\midrule
ELECTRA          & News    & 0.9366 & 0.9673 & 0.9366 & 1.000  & 0.7869 & 35.60 \\
ELECTRA          & Reddit  & \textbf{0.6202} & 0.7103 & \textbf{0.6495} & 0.7851 & 0.6172 & 963.33 \\
ELECTRA          & Twitter & 0.7601 & 0.7898 & 0.7943 & 0.7882 & 0.8372 & 114.32 \\
\midrule
TinyBERT         & News    & 0.9366 & 0.9673 & 0.9366 & 1.000  & \textbf{0.8540} & \textbf{21.81} \\
TinyBERT         & Reddit  & 0.6056 & 0.7041 & 0.6361 & \textbf{0.8032} & 0.6061 & \textbf{620.16} \\
TinyBERT         & Twitter & 0.7553 & 0.8013 & 0.7489 & \textbf{0.8631} & 0.8312 & \textbf{57.68} \\
\bottomrule
\end{tabular}
\end{table*}

\subsection{Platform-Specific Performance Characteristics}

\subsubsection{News Articles: Performance Inflation Due to Class Imbalance}

News content yielded the highest classification accuracy across all models, with average F1-scores reaching 0.9677 ± 0.0011. However, this strong performance is likely inflated by class imbalance introduced through keyword-based data collection. As shown in \autoref{tab:merged-platform-performance}, all models achieved perfect recall (1.0), indicating that most articles were labeled supportive (label 1). This imbalance likely results from sourcing content that favored urban sustainability or coincided with periods of positive policy coverage. Thus, these metrics reflect dataset bias rather than true model capability.

\subsubsection{Reddit: Information Loss Through Summarization}

Reddit posed the greatest classification challenge, with lower average F1-scores (0.7094 ± 0.0034) and accuracy (0.6154 ± 0.0059). These results reflect both the platform’s complexity and information loss from our automated summarization pipeline. As shown in \autoref{tab:merged-platform-performance}, summarizing long-form discussions using Llama-3-8B removed nuanced argumentation and context, which are crucial for accurate stance detection. The close performance across models indicates that the limitation stems from summarization, not model architecture.

\subsubsection{Twitter: Balanced Performance with Moderate Complexity}

Twitter presented moderate classification difficulty, with average F1-scores of 0.8017 ± 0.0071. DistilBERT-base-uncased achieved the highest score (F1: 0.8084), slightly outperforming DistilRoBERTa, suggesting that BERT’s pretraining may be better suited to the informal, abbreviated language typical of Twitter. \autoref{tab:merged-platform-performance} summarizes performance metrics across models. While more interpretable than News, Twitter data presents unique challenges, including compressed arguments, ambiguous hashtags, and concise expression of complex views.

\subsection{Efficiency, Generalization, and Statistical Trends}

Compact transformer models demonstrated strong performance with minimal computational cost. As shown in \autoref{tab:merged-overall-efficiency}, TinyBERT achieved the highest efficiency (3.535 F1 points/sec × 1000), over 60\% higher than the next-best model (ELECTRA), making it well-suited for resource-constrained or real-time applications. Despite aggressive compression, TinyBERT maintained competitive performance (F1: 0.8242), challenging assumptions that compactness implies lower utility.

To evaluate model generalization, we examined the F1-score gap between each model’s best and worst platforms. MiniLM showed the smallest gap (0.2560), indicating the highest cross-platform consistency, followed closely by DistilRoBERTa (0.2566) and ELECTRA (0.2570). In contrast, TinyBERT had the largest gap (0.2632), suggesting greater sensitivity to domain variation. These findings highlight MiniLM’s robustness for generalized applications, while TinyBERT excels in low-latency deployment.

Correlation analysis revealed that accuracy and F1-score were strongly aligned ($r = 0.992$), while training time showed no clear link to performance. All models followed the same platform difficulty pattern: News (best), Twitter (moderate), and Reddit (worst). Pairwise comparisons further emphasized the strength of compressed models. TinyBERT outperformed ELECTRA by 0.0017 F1, while DistilRoBERTa exceeded DistilBERT by only 0.0011—highlighting marginal differences. MiniLM surpassed ELECTRA by 0.0051 and had the best stability overall. These results affirm the practical viability of compressed transformers for efficient, cross-domain sentiment analysis.

\section{Conclusions}

Sentiment analysis has been extensively used by researchers \cite{DasZ21cd, Das15c, Dutta21c}. This study presents the first cross-platform benchmark for sentiment classification on 15-minute city discourse, using data from Twitter, Reddit, and news articles. We evaluated five compressed transformer models (DistilRoBERTa, DistilBERT, MiniLM, ELECTRA-small, and TinyBERT) demonstrating that lightweight models can deliver competitive performance with significantly reduced computational cost. Platform-specific performance varied widely, with F1-scores ranging from 0.7094 (Reddit) to 0.9677 (Online News). The inflated News performance is likely due to class imbalance caused by keyword-based scraping, as all models achieved perfect recall (1.0), suggesting a predominance of supportive articles. These metrics reflect dataset bias rather than model strength.

Reddit posed the greatest challenge due to long, nuanced discussions that were summarized using Llama-3-8B to fit model input constraints. This summarization likely led to semantic loss and reduced classification accuracy. Twitter showed moderate difficulty, offering informal yet stance-rich content, with an average F1 of 0.8017. These findings establish a platform difficulty hierarchy: Reddit $>$ Twitter $>$ News. Thus, model selection must consider platform traits, efficiency, and deployment context. DistilRoBERTa achieved the highest F1 (0.8292), while TinyBERT offered the best efficiency–performance trade-off (F1: 0.8242, training time: 233.22s). MiniLM demonstrated the strongest cross-platform generalization with the smallest F1-gap (0.2560). Correlation analysis confirmed that F1 closely aligns with accuracy ($r = 0.992$) and recall ($r = 0.973$), while training time negatively correlates with AUC ($r = -0.819$). Notably, TinyBERT outperformed ELECTRA by 0.0017 F1 despite aggressive compression, and MiniLM outperformed ELECTRA by 0.0051 while maintaining greater stability. These results affirm the viability of compressed models for scalable sentiment analysis.

However, key limitations remain. Summarization likely removed crucial context, especially on Reddit. Keyword-based scraping caused class imbalance in News. The annotation pipeline, though efficient, lacked human oversight and may reflect LLM biases. All experiments were conducted in a CPU-only environment, limiting evaluation of larger or long-sequence models. Future work should: (1) explore attention-preserving summarization, chunking, or hierarchical transformers for long texts; (2) benchmark larger or instruction-tuned LLMs with few-shot prompts; and (3) extend analysis beyond binary sentiment to multi-label or aspect-based classification, capturing richer views on sustainability, equity, and personal freedom. In summary, compact transformer models, when combined with consistent preprocessing and evaluation, enable efficient and robust cross-platform sentiment classification—offering valuable tools for analyzing polarized public discourse in urban planning contexts.

\bibliographystyle{IEEEtran}
\bibliography{main}

\end{document}